\documentclass[letterpaper, 10 pt, conference]{ieeeconf}  % Comment this line out if you need a4paper

\usepackage{amssymb}
\usepackage{graphicx}
\usepackage{amssymb}
\usepackage[utf8]{inputenc}
\usepackage{url}
\usepackage{float}
\usepackage{amsmath}

\DeclareMathOperator*{\argmin}{arg\,min}
\usepackage{graphicx}
\usepackage{wrapfig}
\usepackage{cite}
\usepackage{caption}
\usepackage{booktabs}
\usepackage{fancyhdr}
\usepackage{lastpage}
\usepackage{multirow}
\usepackage{algorithm}
\usepackage{algpseudocode}

\usepackage[all]{nowidow}
\usepackage[usenames,dvipsnames]{xcolor}
\usepackage{authblk}
\usepackage{lipsum}
\usepackage{subcaption}
\usepackage{indentfirst}
\usepackage[bottom]{footmisc}
\usepackage{titlesec}

\IEEEoverridecommandlockouts                              % This command is only needed if
\title{\LARGE \bf
Multi-Object RANSAC: Efficient Plane Clustering Method in a Clutter
}

\author{Seunghyeon Lim$^{1}$, Youngjae Yoo$^{2}$, Jun Ki Lee $^{3*}$ and Byoung-Tak Zhang$^{1,2,3*}$% <-this % stops a space
\thanks{This work was partly supported by the IITP (2021-0-02068-AIHub/15\%, 2021-0-01343-GSAI/20\%, 2022-0-00951-LBA/25\%, 2022-0-00953-PICA/25\%) and NRF (RS-2023-00274280/15\%) grant funded by the Korean government.}% <-this % stops a space
\thanks{$^{*}$Corresponding authors.}%
\thanks{$^{1}$Interdisciplinary Program in Cognitive Science, Seoul National University, Seoul, Korea}%
\thanks{$^{2}$Department of Computer Science and Engineering, Seoul National University, Seoul, Korea}%
\thanks{$^{3}$AIIS, Seoul National University, Seoul, Korea}%

\\{\tt\small \{shlim, yjyoo\}@bi.snu.ac.kr, junkilee@snu.ac.kr, btzhang@bi.snu.ac.kr}}%

\begin{document}

\maketitle
\thispagestyle{empty}
\pagestyle{empty}

%%%%%%%%%%%%%%%%%%%%%%%%%%%%%%%%%%%%%%%%%%%%%%%%%%%%%%%%%%%%%%%%%%%%%%%%%%%%%%%%
\begin{abstract}

% Image processing for robotic grasping has become a key factor in both computer vision and robotic fields.

In this paper, we propose a novel method for plane clustering specialized in cluttered scenes using an RGB-D camera and validate its effectiveness through robot grasping experiments. Unlike existing methods, which focus on large-scale indoor structures, our approach---Multi-Object RANSAC emphasizes cluttered environments that contain a wide range of objects with different scales. It enhances plane segmentation by generating subplanes in Deep Plane Clustering (DPC) module, which are then merged with the final planes by post-processing. DPC rearranges the point cloud by voting layers to make subplane clusters, trained in a self-supervised manner using pseudo-labels generated from RANSAC. Multi-Object RANSAC demonstrates superior plane instance segmentation performances over other recent RANSAC applications. We conducted an experiment on robot suction-based grasping, comparing our method with vision-based grasping network and RANSAC applications. The results from this real-world scenario showed its remarkable performance surpassing the baseline methods, highlighting its potential for advanced scene understanding and manipulation.

\end{abstract}

%%%%%%%%%%%%%%%%%%%%%%%%%%%%%%%%%%%%%%%%%%%%%%%%%%%%%%%%%%%%%%%%%%%%%%%%%%%%%%%%
\section{INTRODUCTION}
\begin{figure}[t]
\centering
\begin{subfigure}[b]{0.235\textwidth}
\includegraphics[width=\textwidth, height=.75\textwidth]{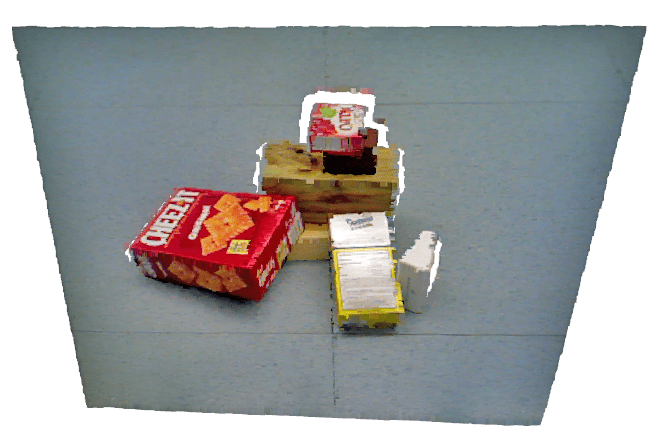}
\caption{The cluttered environment}
\label{fig:fig1a}
\end{subfigure}
\hfill
\begin{subfigure}[b]{0.235\textwidth}
\includegraphics[width=\textwidth, height=.75\textwidth]{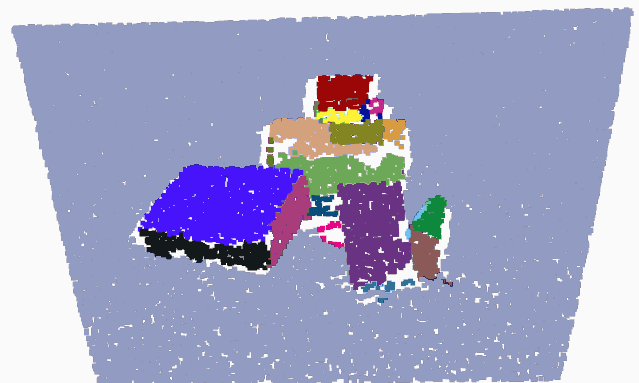}
\caption{Plane clustering}
\label{fig:fig1b}
\end{subfigure}
\caption{A plane clustering result of Multi-Object RANSAC (MO-RANSAC, our method).}
\vspace{-.4cm}
\label{fig:fig1}
\end{figure}

Plane clustering with point clouds plays a pivotal role in finding geometric cues essential for scene understanding in robotics. In particular, plane clustering in cluttered environments (Fig. \ref{fig:fig1a}) is a significant challenge in robotic perception. This challenge is a key enabler for more sophisticated robotic grasping and improved manipulation capabilities. Although there are various methods for clustering planes with a point cloud or depth image \cite{ops2019, pham2016geometrically, hulik2012fast}, these techniques have mainly been designed for scenarios that involve larger objects such as desks, tables, and doors. Our approach---Multi-Object RANSAC (MO-RANSAC), shifts its focus towards cluttered environments encompassing diverse objects of varying scales. By achieving plane instance segmentation within such intricate settings, our method empowers robots to undertake complex grasping and manipulation tasks.

RANSAC, known as RANdom SAmple Consensus, offers a rapid and robust solution for clustering basic shapes, such as planes in our specific case \cite{fischler1981random}. This classical algorithm finds utility across various computer vision tasks, including fitting planes in 3D and lines in 2D. Through an iterative process, RANSAC selectively samples points where the distance from a plane hypothesis falls below a certain threshold (i.e. inlier points). Subsequently, it identifies the optimal parametric model with the greatest consensus among these sampled points. This simple yet efficient approach has gained widespread adoption such as \cite{barath2018graph, brachmann2019neural, barath2019magsac, barath2020magsac++}.

We propose Multi-Object RANSAC which also takes point clouds from the RGB-D data as input. It is composed of two key components: Deep Plane Clustering (DPC) and Post-processing. DPC is trained in a self-supervised manner, generating pseudo-subplane labels through RANSAC. DPC first passes point clouds through a backbone network and subsequently through voting layers \cite{hough2019}, which reorganize the point clouds to effectively group subplane clusters.
Consequently, all points within subplane clusters lie on the same plane. The subsequent step, post-processing, combines multiple clusters that represent the same plane. Fig. \ref{fig:fig1} shows the results of the plane clustering using RGB-D data from the OCID dataset \cite{ocid2019}.

Through this approach, Multi-Object RANSAC learns the geometric attributes of planes. Our experiments have affirmed that using a naive clustering algorithm---an ablated version of our method---would result in the failure to cluster all inlier points located on the same plane. Additionally, such an approach would also lack scalability and adaptability to various environmental factors. On the contrary, Multi-Object RANSAC is trained using point clouds from various environments \cite{ocid2019}, allowing it to effectively mitigate these challenges.

In this paper, we make qualitative and quantitative evaluations of plane clustering. Our experiment is conducted using two datasets: OCID \cite{ocid2019} and OSD \cite{richtsfeld2012segmentation} which encompass RGB-D data captured in various cluttered scenes. To measure the effectiveness of our instance segmentation, we create ground-truth labels by generating plane segments for each object and the background. We compared our approach with recent implementations of RANSAC \cite{barath2018graph, barath2019magsac, barath2020magsac++} and a multi-plane clustering method \cite{ops2019}. In the experiment, Multi-Object RANSAC outperforms the baseline methods across most scenarios. It also shows qualitative improvements in clustering performances on the datasets and real-world cluttered environments.

In real-world scenarios, we performed suction grasp experiments using an UR5 robot arm. The robot was equipped with a RealSense camera at its end effector, receiving RGB-D images. In particular, Multi-Object RANSAC showed a superior grasp accuracy rate compared to existing vision-based suction grasping \cite{suction2021} and plane clustering methods \cite{barath2018graph, barath2019magsac, barath2020magsac++}. Ultimately, the results underline the possibilities of numerous applications for advanced robotic tasks, such as scene understanding and robotic manipulation.

%%% suction grasp experiment를 넣을거면 그 내용도 introduction에 넣어주어야함

\section{Related Works}

\subsection{Plane Clustering}
Plane clustering is of significant importance in the field of robotics. Some studies have employed plane extraction for 3D indoor mapping and localization \cite{pham2016geometrically, kim2018linear}. Oriented point sampling \cite{ops2019} uses an unorganized point cloud with estimated normals to fit planes. However, these methods primarily target expansive indoor environments \cite{song2015sun, Silberman:ECCV12, sturm12iros,handa:etal:ICRA2014}, rendering them unsuitable for complex settings with multiple objects. Many studies\cite{drost2015local, vera2018hough, sommer2020primitect} use the Hough voting to find basic shapes. We developed this strategy using VoteNet voting layers \cite{hough2019}, which represents a recent integration of Hough voting into neural networks.

RANSAC-based shape fitting techniques are versatile and robust approaches with low computational costs. Efficient RANSAC \cite{schnabel2007efficient} can identify multiple geometric shapes by random sampling.
% exhibiting resilience to high noise levels and outliers. 
CC-RANSAC \cite{gallo2011cc} and NCC-RANSAC \cite{qian2014ncc} address RANSAC's limitation in distinguishing between adjacent planes by identifying suitable groups of connected inlier points. 
% Recent advancements take a different route by achieving optimal solutions through energy minimization \cite{barath2018graph} or assigning its own quality scores to each point \cite{barath2019magsac, barath2020magsac++}. 
In addition, some studies\cite{brachmann2019neural, kluger2020consac} use neural networks to efficiently sample inlier points, using self-supervised or supervised methodologies. However, their studies do not emphasize plane fitting or other 3D geometric shapes and focus on computer vision tasks such as estimation of homography and epipolar geometry.

\subsection{Suction Grasp in a Dense Clutter }
Robotic suction grasping is known as an efficient method to handle objects on various scales without damage, widely used in many industrial applications. Recent works focus on methods that directly find grasp points by assigning grasp quality scores in images or 3D spaces.

SuctionNet 1-Billion \cite{suction2021} is an example of a vision-based suction grasping network. It operates by generating heat maps that show the feasibility of suction grasps based on the shapes and physical characteristics of the object. Dex-Net 3.0 \cite{dexnet2018} and Shao et al. \cite{shao2019suction} employ specialized deep learning frameworks designed to extract features to identify graspable suction points.

This paper introduces the application of MO-RANSAC to robotic tasks by suction grasping tasks. Note that MO-RANSAC does not require one to learn grasp points unlike the above methods, but still maintains high accuracy in cluttered environments.

\begin{figure*}[!t]
\centering
\includegraphics[width=\textwidth]{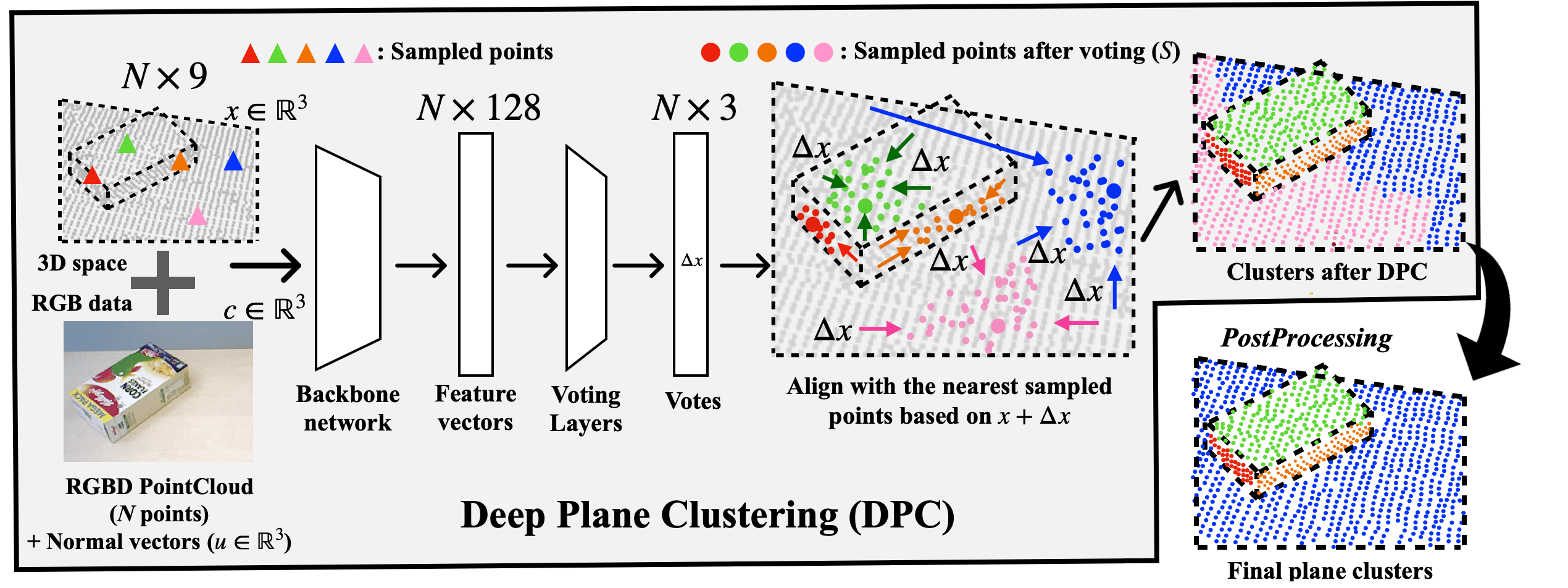}
\caption{\textbf{Overall framework of Multi-Object RANSAC.} Deep Plane Clustering (DPC) takes a 9D point cloud and produces votes in the form of $\{\Delta x\}_{i=1}^N$ s.t. $\Delta x \in \mathbb{R}^3$. Meanwhile, DPC initially samples $K$ points (triangle-shaped points), to individually represent clusters after voting (large circle-shaped points, S). Each point is then associated with the nearest sampled point, resulting in the creation of $K$ subplane clusters. These subplane clusters undergo further refinement through post-processing following DPC.}
\label{fig:overall_framework}
\vspace{-.3cm}
\end{figure*}

\setlength{\textfloatsep}{10pt}

\begin{algorithm}[t]
\caption{Multi-Object RANSAC}
\label{algo:alg1}
\begin{algorithmic}[1]
\Require{$X\! =\! \{x_{i}\}_{i=1}^{N}$, RGB data $\{c_i\}_{i=1}^N$, normals $\{u_i\}_{i=1}^N$  \\\Comment{$x_i, c_i, u_i \in \mathbb{R}^3$}}
\State{Sample $K$ points from $X$ by Farthest Point Sampling.}
\State{Get $\tilde{X}$ by concatenating RGB data and normals to $X$. \\ \Comment{$\tilde{X} = \{\tilde{x}_i\}_{i=1}^{N}$ s.t. $\tilde{x}_i \in \mathbb{R}^9$}}
\State{$\mathcal{F} \leftarrow$ BackboneNetwork($\tilde{X}$) \Comment{$\mathcal{F} \in \mathbb{R}^{N \times 128}$}}
\State{$\Delta X \leftarrow$\: VotingLayers($\mathcal{F}$)
\\ \Comment{$\Delta X = \{\Delta x_i\}_{i=1}^{N}$} s.t. $\Delta x_i \in \mathbb{R}^3$}
\State{Define $P = \{p_i\}_{i=1}^{N}$ s.t. $p_i = x_i + \Delta x_i$.  }
\State{Define $S \subset P$ that belongs to the sampled $K$ points. \\ \Comment{$|S| = K$}}
\State{Align all points to the nearest $p_s$ s.t. $p_s \in S$.  \\\Comment{Making subplane clusters.}}
\For{each cluster}
\State{Apply RANSAC to determine the plane equation and identify inlier points.}
\EndFor
\State{Apply contrastive loss to gather inliers and push against outliers towards the $p_s$ in line 11. \\ \Comment{pseudo-labels from RANSAC in lines 13-15.}}
\State{Post-processing by merging clusters that lie on the same plane during the inference. \\\Comment{See Algorithm 2.}}
\State{\Return{clusters}}
\end{algorithmic}
\end{algorithm}
\section{Methods}

\subsection{Overview}
\vspace{-.1cm}
Alg. \ref{algo:alg1} provides an overview of the Multi-Object RANSAC (MO-RANSAC) procedure. MO-RANSAC takes a 3D point cloud denoted as $X = \{x_i\}_{i=1}^N \text{ s.t. } x_i \in \mathbb{R}^3$ and utilizes Open3D\cite{open3d2018} for normal estimation, determining the principal axis of nearby points selected through KDTreeSearch. We align RGB and depth images with identical resolutions and transform these 2D images into 3D point clouds using the intrinsic parameters of the camera. As a result, an input point cloud is concatenated with RGB data $\{c_i\}_{i=1}^{N}$ and 3D normal vectors $\{u_i\}_{i=1}^{N}$, forming $\tilde{X} = \{\tilde{x}_i\}_{i=1}^N \text{ s.t. } \tilde{x}_i \in \mathbb{R}^9$ (Alg. \ref{algo:alg1} lines 2-4).  It establishes subplane clusters within the Deep Object Clustering (DPC) phase, where all points belonging to the same subplane cluster are situated on the same plane (Alg. \ref{algo:alg1} lines 1-17). However, these subplane clusters may not represent optimal solutions for plane clustering, as it is not guaranteed that multiple subclusters correspond to the same planar surfaces as shown in Fig. \ref{fig:overall_framework} (Clusters after DPC). For this reason, we add the post-processing (Alg. \ref{algo:alg1} lines 18-19) to effectively merge them into final clusters.

\subsection{Deep Plane Clustering}
Deep Plane Clustering (DPC) serves as a self-supervised deep learning framework designed to cluster points that reside on the same plane. This is achieved by taking advantage of the principles of the RANSAC algorithm. 

In this framework, RANSAC is employed individually for each cluster, generating pseudo-labels that distinguish inliers---points close to the plane---from outliers---points farther away.
Fig. \ref{fig:overall_framework} shows the framework of our model. Before using neural networks, an initial step involves Farthest Point Sampling (FPS) on the 3D point cloud to select $K$ points (Alg. \ref{algo:alg1} line 2). After being processed by the network, each of these sampled points will independently form its subplane cluster (Alg. \ref{algo:alg1} lines 9-12). 

The DPC framework comprises a backbone network and voting layers.  Taking inspiration from VoteNet \cite{hough2019}, we have changed the network to fit our self-supervised plane clustering problem. The backbone network propagates the feature vectors for each point. We use PointNet++ \cite{pointnet2017} for the backbone network and produce a 128-dimensional feature vector for each point in our settings.

The voting layers consist of two fully connected linear networks that produce 3-dimensional features for every point. These features are termed votes, since they are subsequently added to each $x_i$ (i.e., $x_i + \Delta x_i$). Each voting layer is followed by a non-linear network, and the final nonlinear network is succeeded by a 1D BatchNorm \cite{ioffe2015batch} layer.
Obtaining $\Delta X = \{\Delta x_i\}_{i=1}^N$ from the voting layers, DPC uses $\{x_i + \Delta x_i\}_{i=1}^{N}$ to make subplane clusters, denoted by $P$ (Alg. \ref{algo:alg1} line 8).

Once the vote is complete, each point $p_i = x_i + \Delta x_i$ is grouped to create a different subplane cluster $\mathcal{C}_k$. Also, let $S$ denote the set of $K$ sampled after voting. All points in $P$ select the cluster with the nearest L2 distance to the "sampled point" among $K$ points (Alg. \ref{algo:alg1} lines 9-12). For better understanding, we re-assign the sampled point after votes $p_s \in S$ in the $k$th cluster as $p^k$. This allows us to describe the process as follows.

\begin{equation}
\text{if}\quad p \in \mathcal{C}_l, \quad \text{then} \quad l = \argmin_{k= 1, ..., K}{\lVert p - p^k \rVert_2}
\vspace{-.1cm}
\label{eq:eq1}
\end{equation}

After aligning the points with the $K$ clusters, we utilize the RANSAC algorithm with the 3D point cloud $X$ to extract plane equations along with the inlier points within each plane (Alg. \ref{algo:alg1} lines 13-15). This step produces sets of inlier points $I_k$, outlier points $O_k$, which serve as pseudolabels, and 3D plane equations $eq_k$ for the $k$th cluster.

We define a self-supervised contrastive loss function that guides DPC to encourage points in $I_k$ to move closer to $p^k$ and points in $O_k$ to move away from $p^k$ (Alg. \ref{algo:alg1} lines 16-17). In this way, the network learns to group all the points belonging to the ground truth subplane clusters. Let the $j$th point in $\mathcal{C}_k$ as $p_{kj}$, then this is accomplished using a simple contrastive loss $L_k$ with the L1 norm for the $k$th subplane cluster:

\begin{equation}
L_k \!=\! \frac{1}{|I_k|}\!\!\sum_{p_{kj} \in I_k}\!\!\!{\lVert p_{kj}\! - \!p^k \rVert} 
\!+\! \frac{1}{|O_k|}\!\!\sum_{p_{kj} \in O_k}\!\!\!{[\alpha\! -\! \lVert p_{kj}\! - \!p^k \rVert]_{+}}
\vspace{-.1cm}
\end{equation}

% \begin{equation}
% \begin{aligned}
% L_k &= \frac{1}{|I_k|}\sum_{p_{kj} \in I_k}{||p_{kj} - p^k||} \\
% &\quad + \frac{1}{|O_k|}\sum_{p_{kj} \in O_k}{[\alpha - ||p_{kj} - p^k||]_{+}}
% \end{aligned}
% \end{equation}

Here, $[a]_{+} = \text{max}(0, a)$ and $\alpha$ is set to 3. The overall loss function $L$ across the $K$ clusters is defined as:

\begin{equation}
L = \frac{1}{K}\sum_{k = 1}^{K}{L_k}
\vspace{-.1cm}
\end{equation}

During training, the cluster number $K$ is randomly selected between 8 and 196 for each prediction.

Furthermore, we have made a slightly different method in the inference process. Unlike the training process, a 9D point cloud $\tilde{X}$ is initially divided into 3 clusters using K-means clustering, which are then separately fed into DPC, in the steps corresponding to lines 3 to 4 of Alg. \ref{algo:alg1}. This allows DPC to concentrate more on points with similar normal vectors and locations.

% TODO: 요거 p 바꾸기
\begin{algorithm}[t]
\caption{Merge Process in Post-processing}
\label{algo:post-processing}
\begin{algorithmic}[1]
\Require{$cluster\_dict$}
\Ensure{$\mathcal{X}_k \!=\! \{x_{k1}, x_{k2}, ..., x_{k(n-1)}, x_{kn}\} \:\text{s.t.}\: |\mathcal{X}_k|\! =\! n$ \\  \Comment{$x_{kj} \!\leftarrow\! \text{the } j\text{th 3D point w/o voting in the } k\text{th cluster}$ }}

% \State{\Comment{ $cluster\_dict$ consists of $cluster\_id$, plane equation and list of points that belong to the same cluster}}
% \State{$cluster\_dict$ = $DOC(points)$}
% \ForAll{clusters with available $cluster\_id$}
%     \State{further clustering with DBSCAN}
%     \State{Update $cluster\_id$}
% \EndFor
\State{$\mathcal{X}_{cluster\_id}, eq_{cluster\_id} \in cluster\_dict$ }
\State{Define center points of  each cluster as $cluster\_center$}
\State{Get KNN graph for $cluster\_center$ as $G$ which consists of a set of vertices (clusters) and edges $(V, E)$ }
% \State{Define $v_a$ as one of the vertices of $G$}
\While{not all vertices of $G$ are visited}
\State{Define an unvisited vertex $\mathcal{X}_a$ that meets $\mathcal{X}_a \in V$}
\For{$\mathcal{X}_b \in \{\mathcal{X}_b|(\mathcal{X}_a,\mathcal{X}_b) \in E\}$}
\State{$m_a = |\{x|x \in \mathcal{X}_a , d(x, eq_b) < \delta\}| / |\mathcal{X}_a|$}
\State{$m_b = |\{x|x \in \mathcal{X}_b , d(x, eq_a) < \delta\}| / |\mathcal{X}_b|$}
\State{$min_d(\mathcal{X}_a, \mathcal{X}_b) \leftarrow \text{min}(x_{ai}, x_{bj})$ for all $i$ and $j$}
\If{$min_d(\mathcal{X}_a, \mathcal{X}_b)\! <\! \beta$ and $\text{max}(m_a, m_b)\! >\! \gamma$}
\State {Merge $\mathcal{X}_a, \mathcal{X}_b$ into single $\mathcal{X}_a$}
\State{Update $cluster\_dict$}
\EndIf
\EndFor
\EndWhile
\end{algorithmic}
\vspace{-.1cm}
\end{algorithm}

\subsection{Post-processing Module}
\vspace{-.1cm}
Once the DPC module has performed a clustering on a set of $N$ points, the post-processing module performs a further refinement by merging clusters that are located on the same planar surface.  It is important to note that post-processing is required only during the inference phase and is not applied during the training. The post-processing module continuously combines two other nodes by the merge process. The details are explained in Alg. \ref{algo:post-processing}.

The merge process itself is a graph-based algorithm whose graph is constructed by establishing connections between nearest neighbors based on the centroids' distances between clusters. In the context of Alg. \ref{algo:post-processing}, a data structure named $cluster\_dict$ is defined. This structure stores cluster identifiers ($cluster\_id$), plane equations ($eq_{cluster\_id}$) obtained in Alg. \ref{algo:alg1} lines 13-15, and the sets of points without voting values associated with these clusters ($\{\mathcal{X}_k\}_{k=1}^K$). After the graph is created successfully, we compare two connected vertices (clusters). 
% by Union \& Find algorithm. 
Two vertices are merged (a) if the minimum distance between points in set A ($\mathcal{X}_a$) and set B ($\mathcal{X}_b$) is less than $\beta$, and (b) if the portion of points in $\mathcal{X}_a$ whose distances from plane equations $eq_b$ fall below $\delta$ exceeds the threshold $\gamma$, and vice versa (as shown in Alg. \ref{algo:post-processing} lines 8-14). We have set $\beta$, $\gamma$, and $\delta$ at 0.2, 0.9, and 0.005, respectively.

At inference, $K$ subplane clusters ($cluster\_dict$) are generated for each of the three large point sets from K-means clustering in DPC. Initially, the merge process is applied to individual sets of $cluster\_dict$ for each $K$, followed by a secondary merge process for all $cluster\_dict$ after the first merge.

The primary bottleneck in terms of computational cost occurs in line 11 of Alg. 2, where we need to compare all points between two clusters in each iteration. Assuming that each cluster has $\mathcal{U}$ nearest neighbors and there are a total of $N$ points, the computational cost becomes $O(\mathcal{U}KN^2)$. 
% Ultimately, this process yields final plane clusters, each representing distinct planar surfaces associated with multiple objects as well as a background.

\section{Experiments}
\subsection{Experiments Setup}
We used the OCID\cite{ocid2019} dataset for training DPC.
% OCID contains RGB-D images of 96 cluttered scenes and 89 objects with various environments.
We randomly sampled the dataset and split it into the train and test sets. MO-RANSAC was trained with the 2,080 scenes and tested with the 300 scenes. We also used the OSD\cite{richtsfeld2012segmentation} dataset for the test dataset, which contains cluttered scenes from table views. We trained MO-RANSAC on an NVIDIA TITAN RTX for 10 epochs with a learning rate of $10^{-5}$ and weight decay of $10^{-5}$. Each point cloud from a single image is downsampled to 32768 points. 

We conducted experiments with three recently developed RANSAC methods, GC-RANSAC \cite{barath2018graph}, MAGSAC \cite{barath2019magsac}, and MAGSAC++ \cite{barath2020magsac++}, as our baselines. We implemented multi-plane clustering iteratively since they made a single model hypothesis per execution. The methods continuously identify a plane hypothesis, extracting inlier points, and then proceed with the same process for the remaining points in the point cloud. We also compared MO-RANSAC with Oriented Point Sampling (OPS) \cite{ops2019}---a multiple plane clustering algorithm with point clouds designed for robotic applications.
% Notably, all these baselines operate on point cloud data as their input.

\subsection{Qualitative Results}
\begin{figure*}[t]
\centering
\includegraphics[width=\textwidth, ] {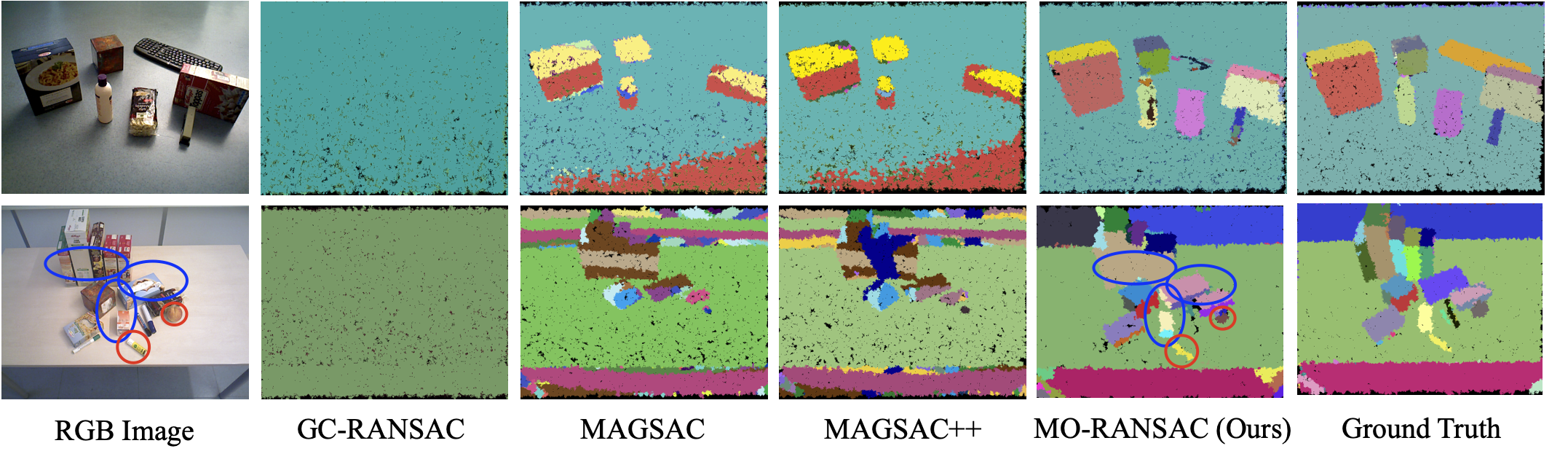}
\caption{Segmentation results compared to plane clustering baselines: red circles represent successful segmentation outcomes for small objects, while blue circles denote segmentation failures for nearby objects.}
\label{fig:fig4}
\vspace{-.1cm}
\end{figure*}
\begin{figure*}[t]
\centering
\includegraphics[width=\textwidth]{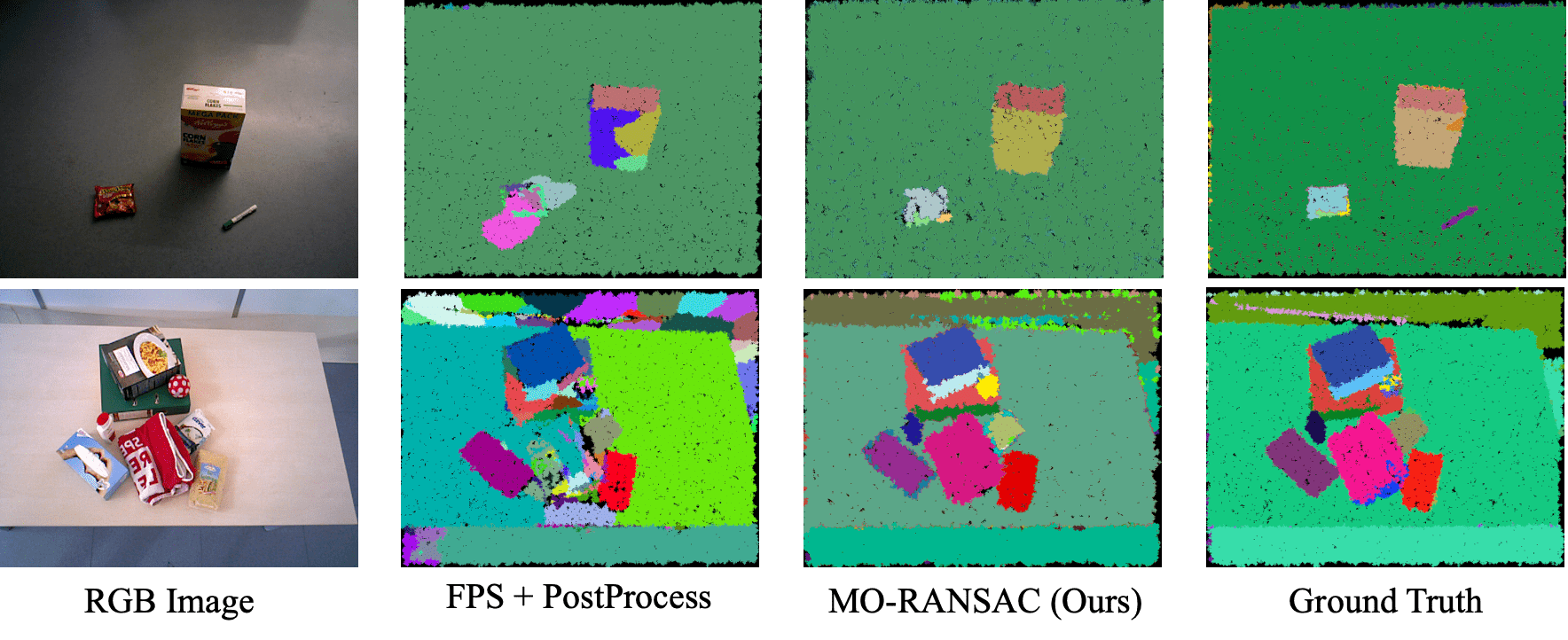}
\caption{Ablation studies of MO-RANSAC and the ground truth.}
\label{fig:fig5}
\vspace{-.3cm}
\end{figure*}

\begin{table*}
\centering
\begin{minipage}{0.27\textwidth}
\centering
\includegraphics[width=\textwidth]{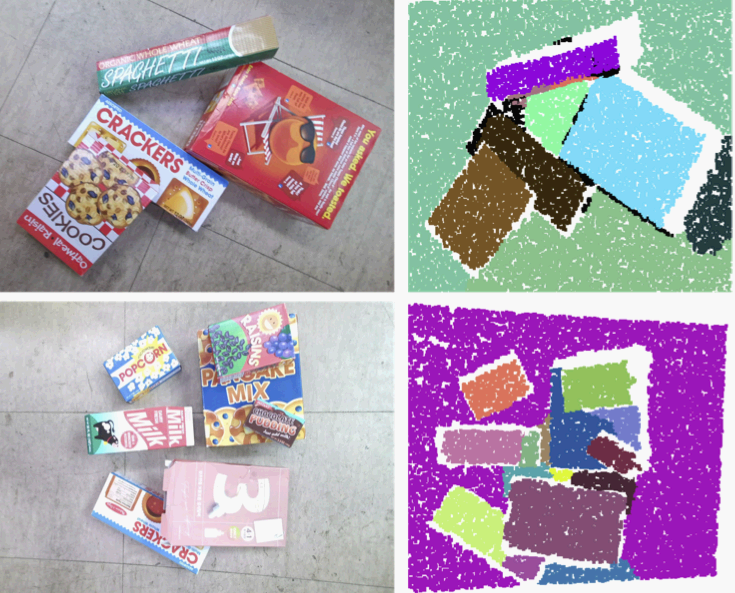}
\captionof{figure}{Real-world examples.}
\label{fig:fig6}
\end{minipage}%
\hspace{.3cm}
\begin{minipage}{0.70\textwidth}
\caption{Segmentation performance comparisons over OCID and OSD datasets.}
\centering
\begin{tabular}{|c|cccccc|}
\hline
\multirow{3}{*}{Method} & \multicolumn{6}{c|}{Datasets}                                                                                                                                                                                \\ \cline{2-7}
& \multicolumn{3}{c|}{OCID \cite{ocid2019}}                                                                                       & \multicolumn{3}{c|}{OSD \cite{richtsfeld2012segmentation}}                                                                   \\ \cline{2-7}
& \multicolumn{1}{c|}{VOI $\downarrow$}            & \multicolumn{1}{c|}{RI}             & \multicolumn{1}{c|}{SC}             & \multicolumn{1}{c|}{VOI $\downarrow$}            & \multicolumn{1}{c|}{RI}             & SC             \\ \hline
GC-RANSAC \cite{barath2018graph}               & \multicolumn{1}{c|}{\textbf{2.584}} & \multicolumn{1}{c|}{0.360}          & \multicolumn{1}{c|}{0.315}          & \multicolumn{1}{c|}{2.181}          & \multicolumn{1}{c|}{0.411}          & 0.389          \\
MAGSAC \cite{barath2019magsac}                 & \multicolumn{1}{c|}{3.004}          & \multicolumn{1}{c|}{0.630}          & \multicolumn{1}{c|}{0.380}          & \multicolumn{1}{c|}{3.392}          & \multicolumn{1}{c|}{0.579}          & 0.332          \\
MAGSAC++ \cite{barath2020magsac++}              & \multicolumn{1}{c|}{2.939}          & \multicolumn{1}{c|}{0.626}          & \multicolumn{1}{c|}{0.384}          & \multicolumn{1}{c|}{3.324}          & \multicolumn{1}{c|}{0.546}          & 0.307          \\
OPS  \cite{ops2019}                   & \multicolumn{1}{c|}{2.989}          & \multicolumn{1}{c|}{0.515}          & \multicolumn{1}{c|}{0.310}          & \multicolumn{1}{c|}{2.594}          & \multicolumn{1}{c|}{0.507}          & 0.372          \\ \hline
MO-RANSAC (Depth)       & \multicolumn{1}{c|}{2.941}          & \multicolumn{1}{c|}{0.738}          & \multicolumn{1}{c|}{0.457}          & \multicolumn{1}{c|}{1.980}          & \multicolumn{1}{c|}{0.871}          & \textbf{0.694} \\
MO-RANSAC (RGB + Depth)  & \multicolumn{1}{c|}{2.845}          & \multicolumn{1}{c|}{\textbf{0.744}} & \multicolumn{1}{c|}{\textbf{0.476}} & \multicolumn{1}{c|}{\textbf{1.974}} & \multicolumn{1}{c|}{\textbf{0.876}} & 0.692          \\ \hline
\end{tabular}

\label{tab:table1}

\end{minipage}
\end{table*}

\begin{figure*}[h]
\centering
\begin{subfigure}{0.22\textwidth}
\includegraphics[width=\textwidth]{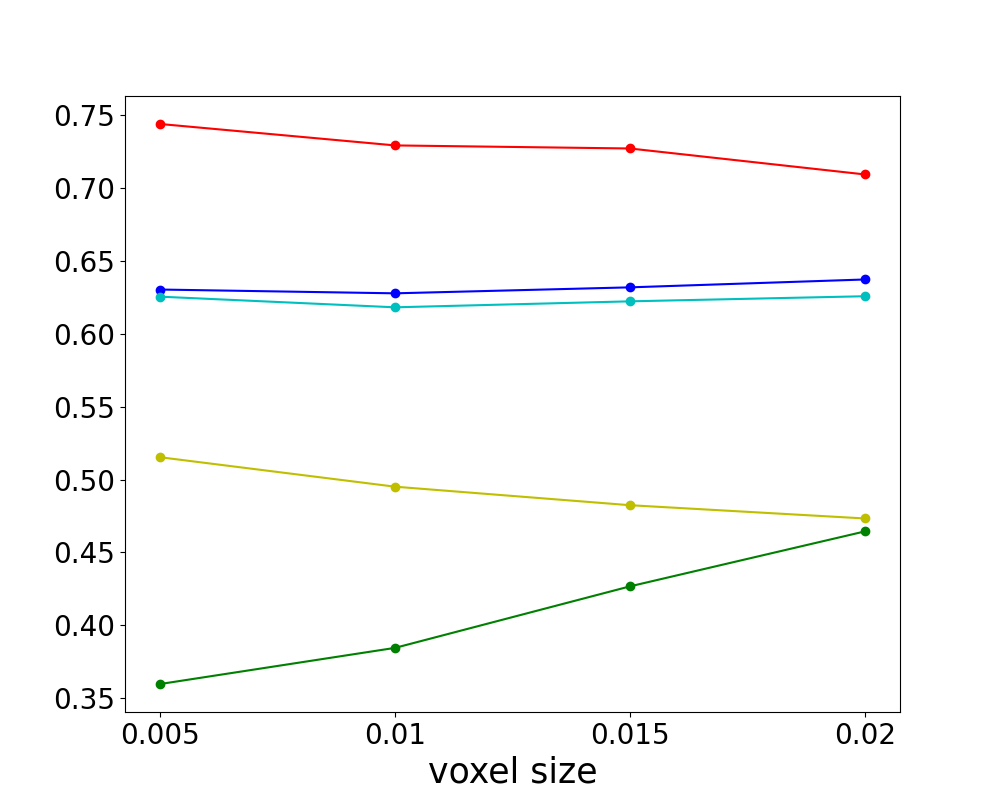}
\caption{OCID - RI}
\label{fig:first}
\end{subfigure}
\hfill
\begin{subfigure}{0.22\textwidth}
\includegraphics[width=\textwidth]{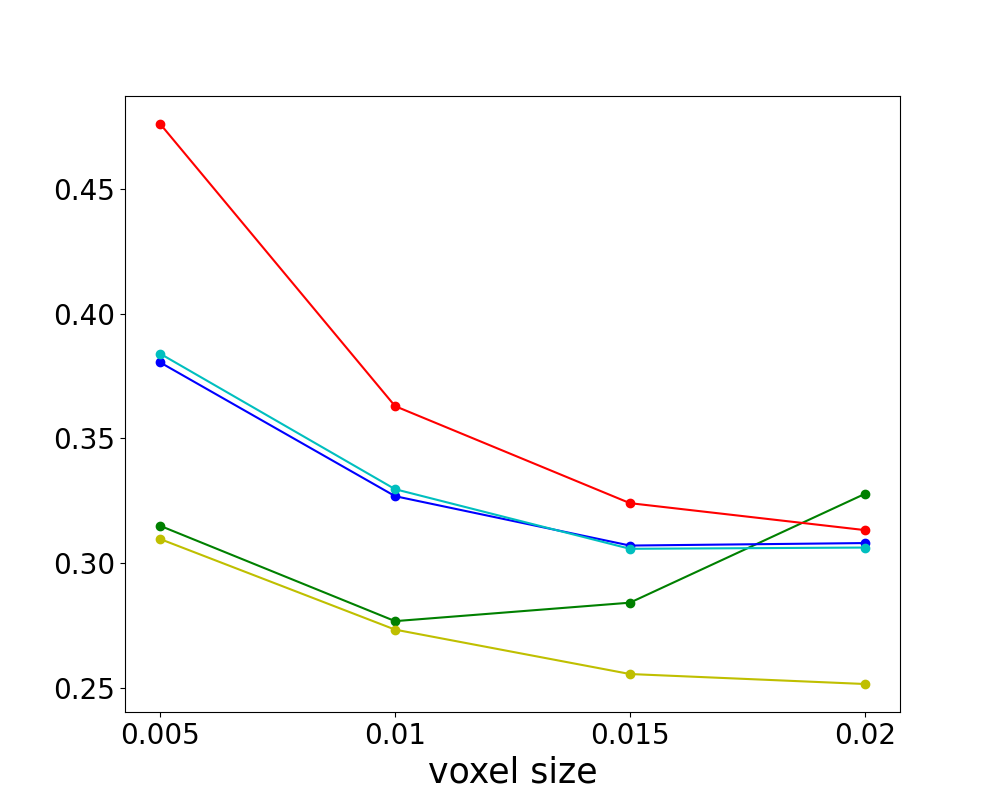}
\caption{OCID - SC}
\label{fig:second}
\end{subfigure}
\hfill
\begin{subfigure}{0.22\textwidth}
\includegraphics[width=\textwidth]{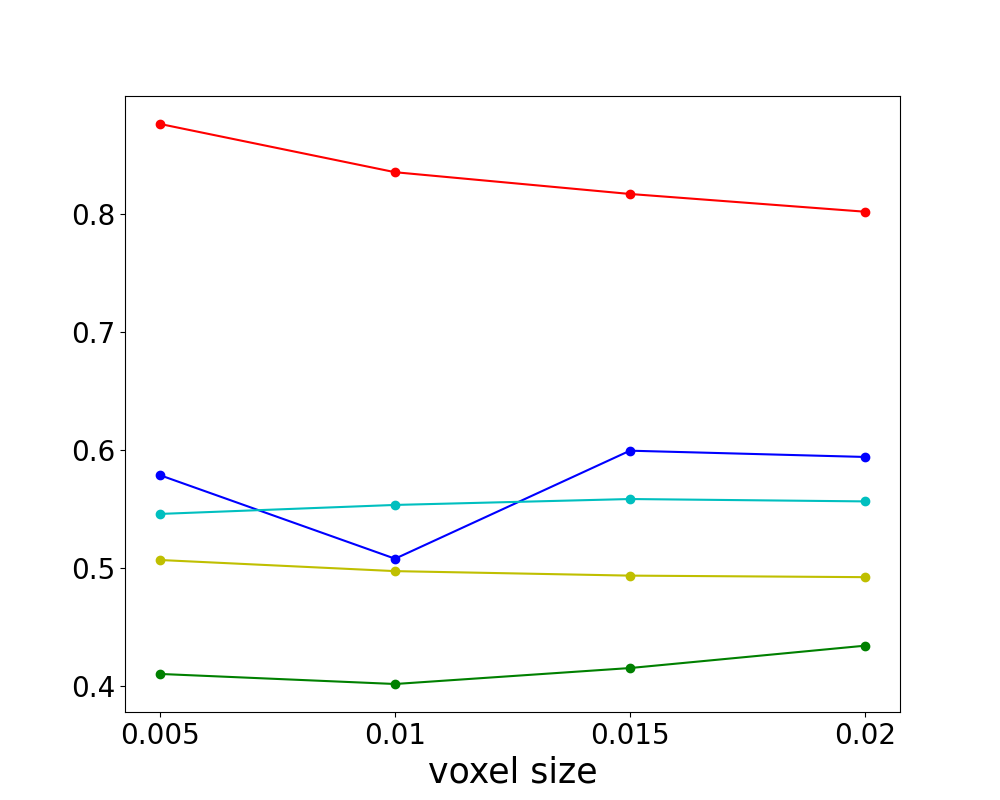}
\caption{OSD - RI}
\label{fig:third}
\end{subfigure}
\hfill
\begin{subfigure}{0.22\textwidth}
\includegraphics[width=\textwidth]{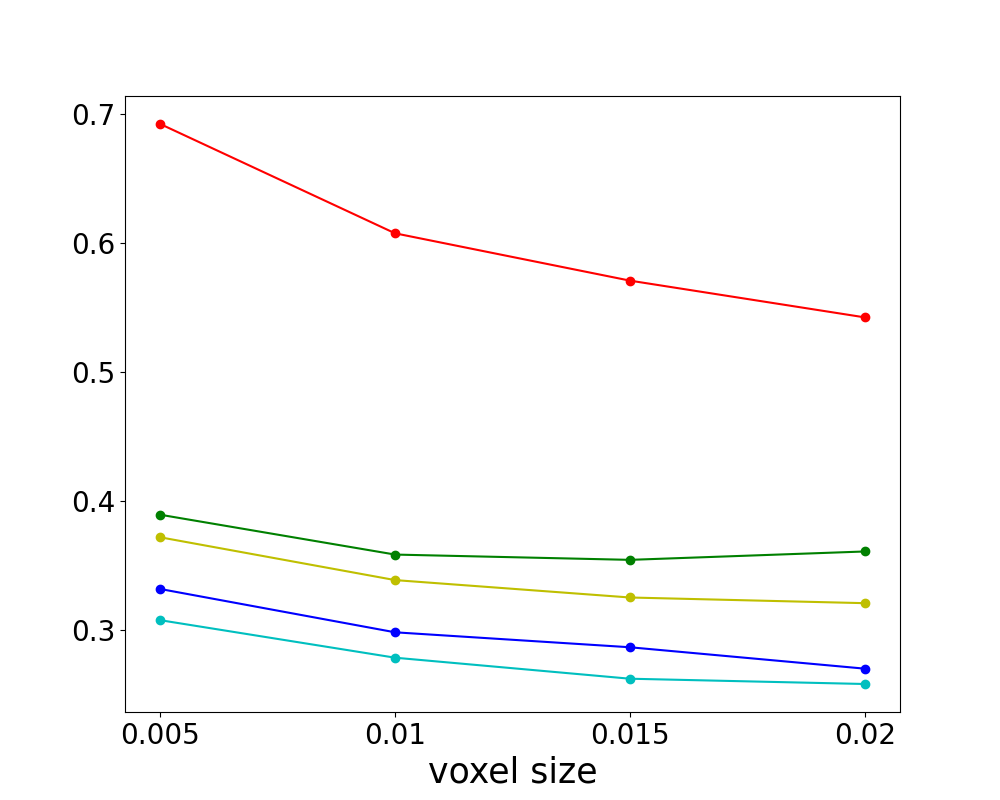}
\caption{OSD - SC}
\label{fig:fourth}
\end{subfigure}
\begin{subfigure}{0.07\textwidth}
\includegraphics[width=\textwidth]{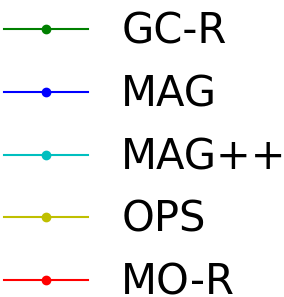}
\label{fig:fourth}
\end{subfigure}
\vspace{-.1cm}
\caption{Performance change of MO-RANSAC based on the \textbf{voxel size} (m) to sample down for evaluation. }
\label{fig:figures_graph}
\vspace{-.5cm}
\end{figure*}

% \begin{table}[h]
% \caption{Rand Index (RI) and loss based on the number of subplane clusters generated by DPC.}
% \begin{center}
% \begin{tabular}{|c|c|c|c|c|}
% \hline
%      & 128   & 256   & 512   & 1024           \\ \hline
% RI   & 0.837 & \textbf{0.839} & 0.829 & 0.830          \\ \hline \hline
% Loss & 0.383 & 0.096 & 0.017 & \textbf{0.003} \\ \hline
% \end{tabular}
% \end{center}
% \end{table}

Fig. \ref{fig:fig4}-\ref{fig:fig6} illustrates the outcomes of plane clustering applied to images from the OCID Dataset and the real-world. Clusters are depicted in distinct colors and projected onto 2D images using camera intrinsic parameters.

Fig. \ref{fig:fig4} shows the comparison of the segmentation results with other RANSAC applications. It is evident that the baseline methods mostly failed to accurately segment surfaces from individual objects. In contrast, MO-RANSAC consistently exhibits superior segmentation performance in all scenarios. Although MO-RANSAC may encounter challenges in distinguishing surfaces amidst closely positioned objects indicated by blue circles, it effectively shows its capabilities by accurately segmenting planar surfaces on smaller objects indicated by red circles.
% Additionally, MO-RANSAC effectively segments both the planetary surface (e.g., boxes) and curved surfaces (e.g., cans), as the points on these curved surfaces align with the same plane hypothesis within a threshold.

In Fig. \ref{fig:fig5}, we made a comparison with an ablated version that makes subclusters without using DPC (FPS + PostProcess). Specifically, it excludes the use of votes ($\Delta x$), opting to create subclusters by $K$ sampled points (Alg. \ref{algo:alg1} line 11) based solely on the $x$ values instead of $p$ (Equation \ref{eq:eq1}). The ablated version produces less refined clustering outcomes, as the points within the same subcluster do not lie on a common plane. This poses a significant risk during the subsequent merge processes in the post-processing.

Additionally, experiments were conducted on real-world objects. Fig. \ref{fig:fig6} shows examples of plane clustering outcomes. We conducted a suction grasping experiment on these objects. The details are given in Section IV.D.

\subsection{Plane Instance Segmentation Result}

We conducted an instance segmentation experiment on the planar clustering results, utilizing the experimental settings detailed in Section IV.A. 
% We created ground-truth labels by extracting pixel information from the datasets. 
% We sampled 12 points using Farthest Point Sampling (FPS) per object to ensure coverage across all planar surfaces of the object and clustering with the merge process (Alg. \ref{algo:post-processing}).
Consistent with previous studies, we employed Variation of Information (VOI), Rand Index (RI), and Segmentation Covering (SC) as evaluation metrics \cite{liu2019planercnn, xie2022planarrecon}.
For evaluation, we transformed the point cloud into voxels with a voxel size of 0.5cm. The experimental results are given in Table \ref{tab:table1}. MO-RANSAC outperforms other clustering algorithms in most scenarios. 
Note that while GC-RANSAC exhibits a relatively high VOI, it shows poor clustering performance as in Fig. \ref{fig:fig4}. This is because it tends to group all the points into a single cluster, unlike other methods which tend to produce multiple smaller false-positive clusters.
We also conducted a comparison involving MO-RANSAC using solely depth information as input (MO-RANSAC (Depth)). While MO-RANSAC with RGB-D data exhibits slight improvements in performance except for SC on the OSD datasets, RGB inputs provide additional visual features that aid in comprehending the object's geometric attributes.

Figure \ref{fig:figures_graph} shows the variations in RI and SC based on the voxel sizes set for evaluation. Here, it becomes evident that MO-RANSAC shows significant enhancements in plane segmentation for smaller voxel sizes, whereas other methods display relatively weak correlations. As smaller voxel sizes encapsulate more intricate object geometric information, we can conclude that MO-RANSAC exhibits robustness in cluttered environments with smaller objects than the others.

\subsection{Suction Grasp Performance in the Real World}

\textbf{Baselines}
We conducted two distinct experiments to evaluate the effectiveness of our suction grasping approach. The first experiment compared MO-RANSAC with the plane clustering baselines (Table \ref{tab:tab2}). In the second experiment, we pit MO-RANSAC against the SuctionNet Baseline \cite{suction2021}, a vision-based suction grasping network that infers graspable points from an RGB-D camera (Table \ref{tab:tab3}). By conducting these experiments using different baseline methods, we aimed to show the usefulness of MO-RANSAC in robotic grasping.

\textbf{Setup}
We use an UR5 robot arm equipped with ROBOTIQ one-cup suction gripper on its end effector. An Intel Realsense D435 RGB-D camera was also affixed to the end effector, and objects were on a flat table within the camera's observable range.
% The control of the robot was facilitated through the URX Python API, while sensor data was obtained from the camera using the ROS interface.
We collected 21 objects from the YCB \cite{7254318}, HOPE \cite{tyree2022hope} datasets, and the local grocery items. To ensure fairness, the experiment in Table \ref{tab:tab3} exclusively employed 6 objects (cracker box, tomato soup can, sugar box, mug, cup, mustard), which were the same as or similar to those used for training the SuctionNet Baseline \cite{suction2021}.

\textbf{Rules} Each experiment consists of 10 rounds. In the experiment in Table \ref{tab:tab2}, each round includes 7 objects: 5 with planar surfaces (e.g., boxes) and 2 with curved surfaces (e.g., cans). In the other experiment (Table \ref{tab:tab3}), 6 objects are present in every round for the experiment. The objective is to successfully grasp and relocate all objects to a predefined location.
% We followed the detailed rules from \cite{vgn2020}.
A round ends when the robot effectively moves all objects to the predefined location. If the robot cannot grasp an object twice consecutively, the object is manually removed and the round continues.

\textbf{Grasp Point Inference} To determine the grasp point, the robot selects the centroid of the cluster farthest from the floor underneath the clustered objects.

\textbf{Evaluation Metrics} We utilize two key metrics: \textbf{success rate} and \textbf{clearance rate} \cite{vgn2020}. The success rate is the ratio of successful grasps to the total attempts.
% ---variations in the number of attempts among methods are due to different trials for successfully picking up individual objects.
The clearance rate represents the proportion of objects successfully cleared by robotic grasping. 

\textbf{Results} MO-RANSAC demonstrates significant enhancements in suction-based grasping in both experiments. In Table \ref{tab:tab2}, MO-RANSAC achieves 16\% higher success rates for planar surface objects and 7\% higher for curved surface objects compared to MAGSAC \cite{barath2019magsac}, which showed the most favorable baseline performance. In particular, the comparative performance of suction grasp is closely correlated with the overall instance segmentation results in Fig. \ref{fig:fig4} and Table \ref{tab:table1}. This emphasizes the critical role of plane instance segmentation in the effectiveness of robotic grasping.

For the experiment in Table \ref{tab:tab3}, MO-RANSAC shows a 23\% increase in success rate and a 26\% increase in clearance rate.
SuctionNet 1-Billion shows a poor grasp accuracy rate due to the mismatch between the experimental environment and the training dataset conditions. This discrepancy exists even though the method uses objects from their list that reported superior performance in their paper \cite{suction2021}. From this perspective, MO-RANSAC emerges as a possible substitute for robotic grasp networks, capable of adapting to various cluttered environments.

\begin{table}[t]
\captionsetup{justification=centering}  % Center-align the caption
\caption{Comparisons of suction grasp accuracy with other plane clustering methods. (S: Success, C: Clearance)}
\vspace{-.3cm}
\begin{center}

\label{tab:tab2}

\begin{tabular}{|c|cc|cc|cc|}

\hline
\multirow{2}{*}{Method} & \multicolumn{2}{c|}{Plane}                                     & \multicolumn{2}{c|}{Curved}                                    & \multicolumn{2}{c|}{Overall}                                   \\ \cline{2-7}
& \multicolumn{1}{c|}{S} & C                       & \multicolumn{1}{c|}{S} & C                       & \multicolumn{1}{c|}{S} & C                       \\ \hline

GC-RANSAC \cite{barath2018graph}             & \multicolumn{1}{c|}{0.302}   & 0.520                           & \multicolumn{1}{c|}{0.452}   & 0.700                           & \multicolumn{1}{c|}{0.342}   & 0.571                           \\
MAGSAC  \cite{barath2019magsac}               & \multicolumn{1}{c|}{0.561}   & 0.740                           & \multicolumn{1}{c|}{0.464}   & 0.650                           & \multicolumn{1}{c|}{0.532}   & 0.714                           \\
MAGSAC++  \cite{barath2020magsac++}             & \multicolumn{1}{c|}{0.551}   & 0.760                           & \multicolumn{1}{c|}{0.364}   & 0.600                           & \multicolumn{1}{c|}{0.490}   & 0.714                           \\ \hline
MO-RANSAC              & \multicolumn{1}{c|}{\textbf{0.717}}   & \textbf{0.860} & \multicolumn{1}{c|}{\textbf{0.536}}   & \textbf{0.750} & \multicolumn{1}{c|}{\textbf{0.659}}   & \textbf{0.829} \\ \hline
\end{tabular}

\end{center}
\vspace{-.3cm}
\end{table}

\begin{table}[t]
\captionsetup{justification=centering}  % Center-align the caption
\caption{Comparison of suction grasp accuracy with a vision-based suction grasping method.}
\vspace{-.3cm}
\begin{center}

\label{tab:tab3}
\begin{tabular}{ |c|c|c|}
\hline
{Method}  & Success          & Clearance       \\ \hline
SuctionNet Baseline \cite{suction2021} & 0.248 (26 / 105) & 0.433 (26 / 60)  \\ \hline
MO-RANSAC            & \textbf{0.482 (42 / 87)}  & \textbf{0.700 (42 / 60)} \\ \hline
\end{tabular}
\end{center}
\vspace{-.3cm}
\end{table}
\section{CONCLUSIONS}
\vspace{-.2cm}
This paper introduces MO-RANSAC, a novel plane clustering technique specially designed for cluttered environments involving multiple objects, leveraging data from an RGB-D camera. MO-RANSAC efficiently rearranges points via voting layers for plane clustering. MO-RANSAC excels in complex plane clustering scenarios and shows promise for real-world robotics applications, including suction grasping.
% In our experimental evaluation, MO-RANSAC demonstrates superior performance in plane clustering within complex environments. 
% Finally, the results of suction grasp experiments suggest that MO-RANSAC holds promise as an applicable solution to real-world robotic tasks.

\section*{ACKNOWLEDGMENT}
\vspace{-.2cm}
We extend our heartfelt gratitude to Jae-in Kim and Hee-bin Yoo for their invaluable insights and help with the revision of this manuscript.

%%%%%%%%%%%%%%%%%%%%%%%%%%%%%%%%%%%%%%%%%%%%%%%%%%%%%%%%%%%%%%%%%%%%%%%%%%%%%%%%

% References are important to the reader; therefore, each citation must be complete and correct. If at all possible, references should be commonly available publications.

% \section*{Bibliography}
% \newpage
\bibliographystyle{IEEEtran}
\bibliography{lsh2023}

\end{document}